\documentclass{article}
\usepackage{spconf,amsmath,graphicx}
\usepackage{xcolor}
\usepackage{multirow}
\usepackage{booktabs}

\title{Text2Avatar: Text to 3D Human Avatar Generation with Codebook-Driven Body Controllable Attribute}

\name{
Chaoqun Gong$^{1 \dagger}$, Yuqin Dai$^{2 \dagger}$, Ronghui Li$^{1}$, Achun Bao$^{1}$, Jun Li$^{2}$, Jian Yang$^{2}$,
Yachao Zhang$^{1 \star}$, Xiu Li$^{1 \star}$
\thanks{${\dagger}$ These authors contributed equally to this work.}  
\thanks{${\star}$ Corresponding authors:  yachaozhang@sz.tsinghua.edu.cn,\\li.xiu@sz.tsinghua.edu.cn } 
\thanks{© 2024 IEEE.  Personal use of this material is permitted.  Permission from IEEE must be obtained for all other uses, in any current or future media, including reprinting/republishing this material for advertising or promotional purposes, creating new collective works, for resale or redistribution to servers or lists, or reuse of any copyrighted component of this work in other works.}
}

\address{$^1$Shenzhen International Graduate School, Tsinghua University, China\\ 
$^2$School of Computer Science and Engineering, Nanjing University of Science and Technology, China} 

\begin{document}
\maketitle
\begin{abstract} 
Generating 3D human models directly from text helps reduce the cost and time of character modeling. 
However, achieving multi-attribute controllable and realistic 3D human avatar generation is still challenging due to feature coupling and the scarcity of realistic 3D human avatar datasets.
To address these issues, we propose Text2Avatar, which can generate realistic-style 3D avatars based on the coupled text prompts. 
Text2Avatar leverages a discrete codebook as an intermediate feature to establish a connection between text and avatars, enabling the disentanglement of features. 
Furthermore, to alleviate the scarcity of realistic style 3D human avatar data, we utilize a pre-trained unconditional 3D human avatar generation model to obtain a large amount of 3D avatar pseudo data, which allows Text2Avatar to achieve realistic style generation. 
Experimental results demonstrate that our method can generate realistic 3D avatars from coupled textual data, which is challenging for other existing methods in this field.  
\end{abstract}
\section{Introduction}
\label{sec:intro}
\label{sec:Method}
3D human body modeling has wide-ranging application prospects in film production, video games, human-machine interaction, and content creation. Traditional 3D human body modeling is a complex and costly process, which can take thousands of hours to produce modeling products to meet requirements. Consequently, the utilization of text prompts in cross-modal 3D avatar generation frameworks has emerged as a practical and accessible modeling method with lower entry barriers. 

There have been several instances \cite{avatarclip,dreamavatar} which can generate reasonably matching 3D avatars using prompt words in recent years. To achieve better controllability, some research \cite{clipnerf, clipmesh} enables manipulating NeRF \cite{mildenhall2020nerf} using either a short text prompt or an exemplar image. TeCH \cite{tech} achieves 2D-to-3D human body reconstruction by using coupled text as assistance. However, research focused on generating 3D human bodies using prompt words solely is still scarce. Notably, to the best of our knowledge, due to the absence of realistic-style 3D datasets and generation resolution limitations, all of the current text-to-3D avatar generators without additional information can only produce anime-style results. 

Moreover, it is difficult to decouple the generator's latent space, therefore simultaneously satisfying multiple human attributes in a single generated result is challenging. 
StyleFlow \cite{styleflow} enables decoupled face editing by modifying human face attributes through a reverse inference process. InterFaceGAN \cite{interfacegan} achieves multi-attribute face control by altering the projection direction of the vectors in the subspace latent space. However, existing research has primarily focused on face editing, with limited work on human body decoupling editing due to the more complex spatial structure and the scarcity of datasets,

\begin{figure*}[tbp]
  \centering
  \mbox{} \hfill
  \hfill
  \includegraphics[width=1.0\linewidth]{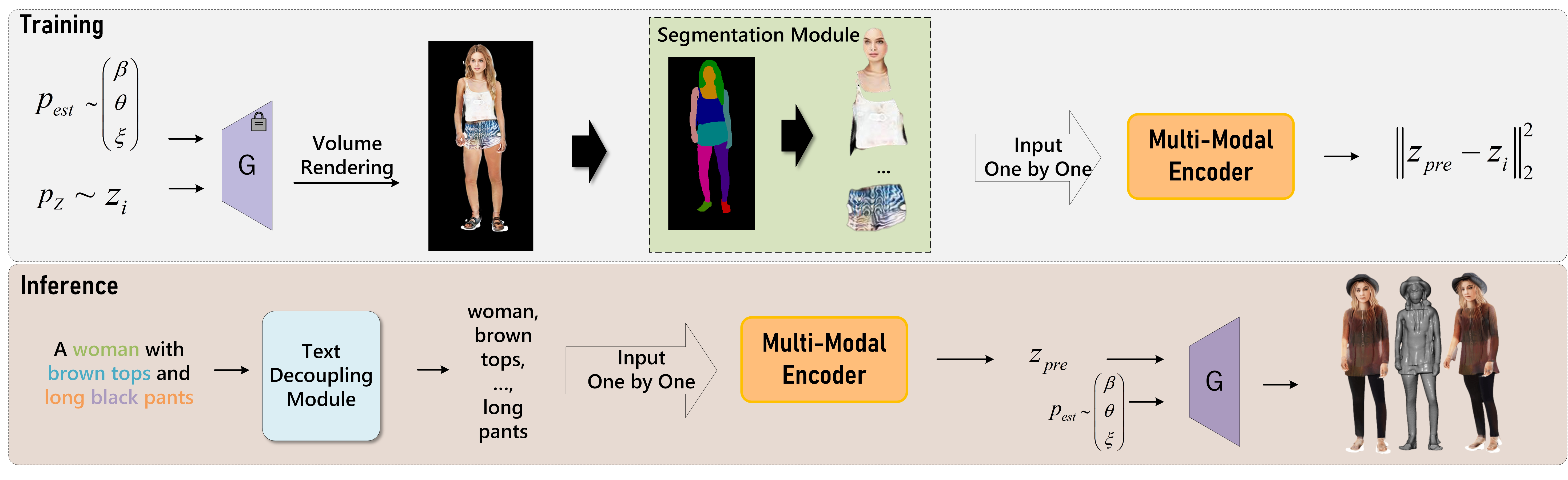}
  \hfill \mbox{}
\vspace{-9mm}
  \caption{\label{fig:framework_sim1}%
           Framework of Text2Avatar. }
\vspace{-3mm}
\end{figure*}

In this paper, we propose a novel framework, named Text2Avatar, which can generate 3D avatars from multi-attribute prompts containing human clothing information. Different from the generation of objects \cite{ lin2023consistent123}, the generation of avatars focuses more on the rationality and controllability of body elements. Unlike \cite{tech}, we only rely on textual prompts without the need for additional image inputs, thereby possessing higher levels of difficulty and a broader range of application prospects. 
To realize cross-modal generation, we proposed the Multi-Modal Encoder, which can be used as a plugin to assist in unconditionally generating models for textual cross-modal tasks. Inspired by prior works \cite{pix2nerf, hong2023evad, jiang2022text2human}, we employ discrete attribute codes to express the 3D human body, realizing decoupled representation. By using the existing cross-modal model CLIP(Contrastive Language-Image Pre-Training) \cite{clip}, which provides a paired semantic-consistent text-image encoder, we are able to encode text/image features into the discrete codebook. The codebook contains the human body feature and serves as a mediator to obtain the matching latent code, which controls the 3D avatar generation. To achieve high-accuracy attribute matching encoding, we employed a segmentation module \cite{seg} to support the CLIP model. In addition to the segmentation module, we utilized inherent image information(e.g., RGB) to facilitate the matching process. 

We extensively evaluate our proposed Text2Avatar and demonstrate that when presented with coupled textual prompts, our framework can generate high-quality 3D clothed avatars that satisfy complex attribute requirements.

\begin{figure}[tbp]
  \centering
  \mbox{} \hfill
  \hfill
  \includegraphics[width=1.0\linewidth]{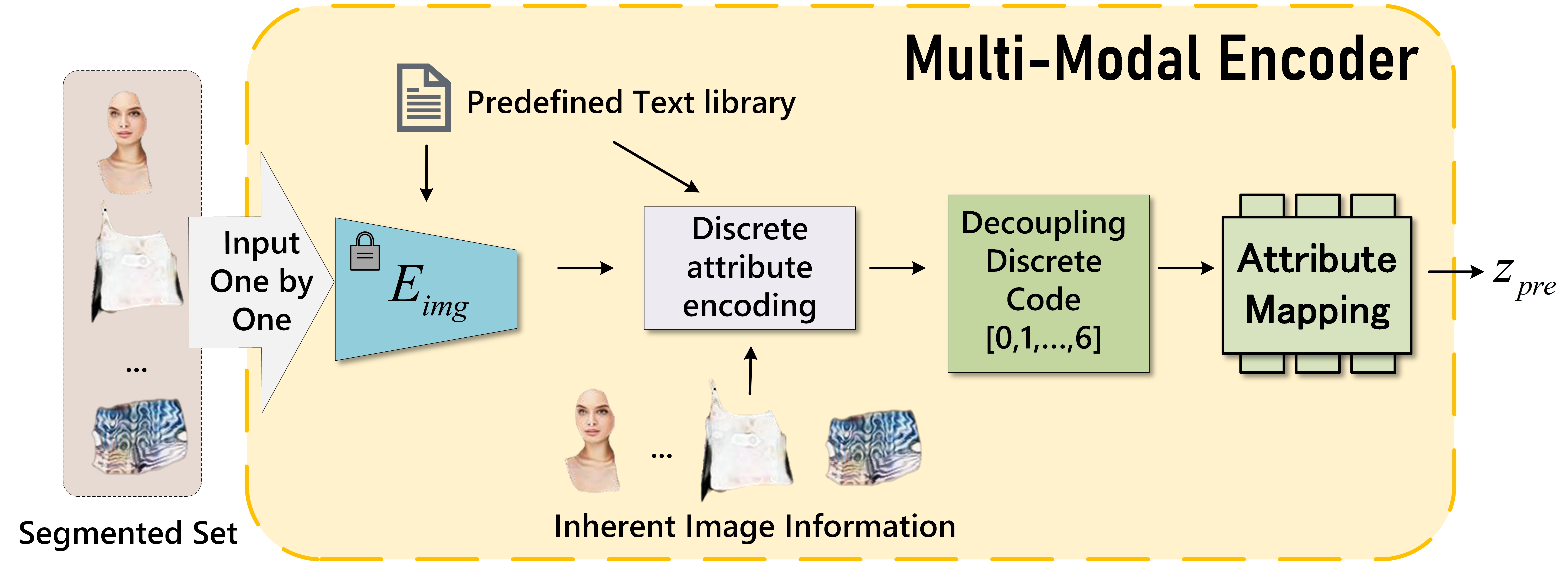}
  \hfill \mbox{}
  \vspace{-8mm}
  \caption{\label{fig:enc}%
           Multi-Modal Encoder. }
\vspace{-4mm}
\end{figure}

\section{Method}
\subsection{Structural Composition}
Overall, our structure consists of three neural networks: the 3D-aware GAN generator $G(\cdot)$, segmentation module $Seg(\cdot)$ the multi-modal encoder $E(\cdot)$. The input of the generator $G(\cdot)$ includes camera parameters distribution $p_{\text{text}}$ and the latent code $\boldsymbol{z} \sim p_z$. The $p_{\text{text}}$ includes the shape parameter $\beta$, the pose parameter $\theta$, and the perspective attitude $\xi$.
The multi-modal encoder $E(\cdot)$ consists of cross-modal text encoder $E_{\text{text}}(\cdot)$ and image encoder $E_{\text{text}}(\cdot)$, an attribute mapping network $M(\cdot)$, and a predefined text library$\{t_{\text{predefined}_{ij}}|i = 1,...,n, j = 1,...,n^i_{attr}$, where $n$ is the number of the attribute, and $n_{attr,i}$ represents the number of predefined categories for the $i$-th attribute. Our Framework is shown in Figure \ref{fig:framework_sim1}.

\begin{table*}[h!]
  \begin{center}
    \caption{Comparison of different baselines w.r.t. attribute accuracy and R-Precision.\label{table:com}}
    \setlength{\tabcolsep}{3pt} 
    \begin{tabular}{l|ccccccc|cc} 
    \toprule
      \multirow{2}{*}{\textbf{Methods}} & \multicolumn{7}{c|}{\textbf{Attribute Accuracy}} & \multicolumn{2}{c}{\textbf{R-Precision}}\\
      & Gender & Sleeve-length & Top-color & Top-type & Pants-length & Pants-color & Pants-type & ViT-B/32 & ViT-L/14\\
      \hline
      DreamFusion & 1.00 & - & - & - & - & - & - & 74.71 & 79.64 \\
      3DFuse & 1.00 & 0.30 & 0.65 & 0.20 & 0.55 & 0.40 & 0.15 & 77.83 & 82.76\\
      AvatarCLIP & 1.00 & - & 0.60 & - & - & 0.40 & - & 76.66 & 81.15 \\
      Text2Avatar & \textbf{1.00} & \textbf{1.00} & \textbf{0.80} & \textbf{0.55} & \textbf{0.85} & \textbf{0.90} & \textbf{0.60} & \textbf{78.52} & \textbf{83.30}\\
    \bottomrule 
    \end{tabular}
  \end{center}
  \vspace{-5mm}
\end{table*}

It should be emphasized that the text library is primarily designed to complement the encoders provided by CLIP \cite{clip} for attribute matching. Due to the strong data support of CLIP, this text library can theoretically be easily expanded without the need for additional training steps. The segmentation module \cite{seg} is employed to convert the local information of an image into global information in order to enhance the performance of the CLIP model. Motion generation methods \cite{li2023finedance} can be applied to the generation of $\beta$ so that the avatar presents a variety of poses or dances.

\subsection{Multi-Modal Encoder}
\label{sec:enc}

The Multi-modal Encoder can serve as a plugin to assist unconditional generation models in 
textual cross-modal tasks. The Framework is shown as Figure \ref{fig:enc}. 
Given a 2D human body rendering segmented set $\{I^i_{\text{seg}}|i = 1,...,n\}$, we utilize CLIP to extract the decoupled attribute feature.
Specifically, by utilizing pre-trained image encoder $E_{\text{img}}(\cdot)$ and text encoder $E_{\text{text}}(\cdot)$, which can encode visual and textual information into paired features, we identify the most relevant textual description for each $I^i_{\text{seg}}$ within the given text library $\{t^{ij}_{\text{predefined}}|i = 1,...,n, j = 1,...,n^i_{attr}\}$. We encode both the textual and visual content, and compute their cosine similarity to select the best matching pair. The index of the $t^j_{\text{predefined}}$ is set to be the attribute value $a^i$. The formula for discrete attribute encoding is as follows:
\begin{equation}
    \begin{aligned}
    a^i = \arg\max_j \frac{E_{\text{img}}(I^i_{\text{seg}}) \cdot E^T_{\text{text}}(t^{ij}_{\text{predefined}})}{\left\lVert E_{\text{img}}(I^i_{\text{seg}}) \right\rVert \left\lVert E_{\text{text}}(t^{ij}_{\text{predefined}}) \right\rVert}
    \end{aligned}
\end{equation}
The attribute mapping network will then utilize the aforementioned codebook to obtain the corresponding $z_{\text{gen}}$, enabling control over unconditional generative models.

\subsection{Training Setup}
For the training of generators and discriminators, we follow the training methods of Hong et al \cite{hong2023evad}.

We train an attribute mapping network based on MLP, which is mainly used to map the image human-attribute space to the latent space of the generative model. Unlike the video generation \cite{ma2023follow}, avatar appearance is affected by the latent code $z$. Specifically, we first use the generative model to obtain paired latent variables $z_{\text{gen}}$ and the corresponding 3D avatar. We use volume rendering to obtain the corresponding image $I_{\text{gen}}$. Then, utilizing the image encoder $E_{\text{img}}(\cdot)$ provided by CLIP and the segmentation module \cite{seg}, we encode the image $I_{\text{gen}}$ with a carefully-designed text library $\{t^{ij}_{\text{predefined}}|i = 1,...,n, j = 1,...,n^i_{attr}\}$ into discrete codebook. The codebook corresponds to the attribute information with the corresponding number in the text library. After that, we use the mapping network to map the codebook to the latent space of the generative model, obtaining the predicted latent code $z_{\text{pre}}$. We compare this $z_{\text{pre}}$ with the true $z_{\text{gen}}$, and optimize the mapping network using MSE loss. The objective function $L(\phi)$ is shown below: 
\begin{equation}
L(\phi) = \min _\phi \frac{1}{N} \sum_{i=1}^N\left\|z^i_{\text{pre}}-z^i_{\text{gen}}\right\|^2,
\end{equation}
where $\phi$ represents the parameters of the mapping network, and $N$ denotes the number of training instances.

\begin{figure}[tbp]
  \centering
  \mbox{} \hfill
  \hfill
  \includegraphics[width=1.0\linewidth]{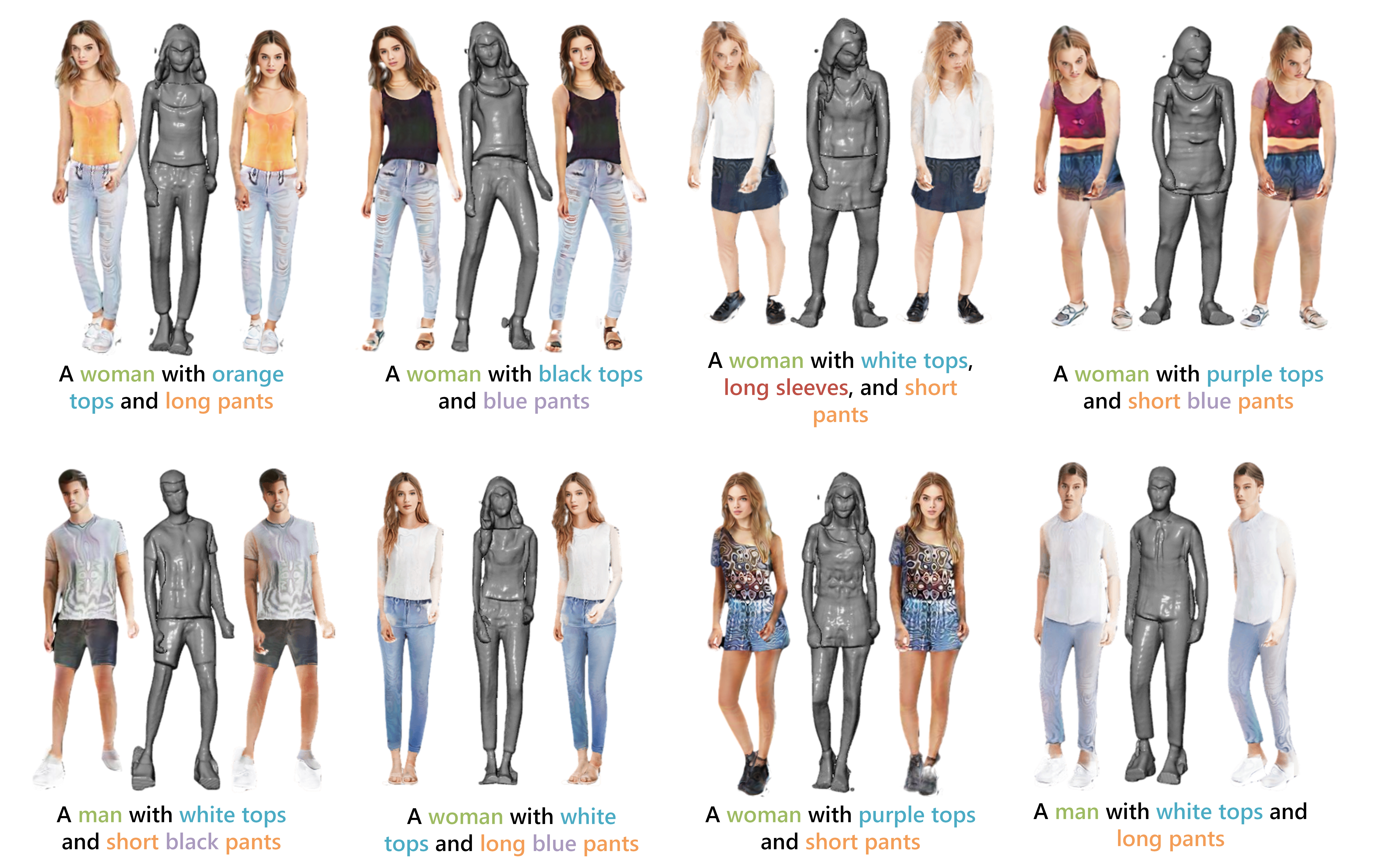}
  \hfill \mbox{}
  \vspace{-7mm}
  \caption{\label{fig:gen}%
          Generation results from coupled textual prompts. }
\vspace{-5mm}
\end{figure}

\subsection{Decoupling Control Generation}
During the model inference phase for 3D avatar generation, the model takes as input text describing human attributes. The input text is first decoupled into various human attributes. These attributes are then fed into the multi-modal encoder that utilizes CLIP for matching with a text library to obtain human attribute code $a$. Then the attribute code  are encoded into latent code $z$ using the attribute mapping network $M(\cdot)$ trained during the training phase.

The latent code $z$, along with shape parameters $\beta$ and pose parameters $\theta$ and camera viewpoint $\xi$, are input into a generator to produce an image of the avatar from a certain perspective during rendering. The mesh of avatar can be obtained by using the offsets inferred by a special NeRF within the generator.

\section{Experiment}
\label{sec:Experiment}
\vspace{-1mm}
\subsection{Implementation Details}
\vspace{-2mm}
We implemented code using pytorch on one NVIDIA RTX 3090 GPU. The training of the generator and discriminator follows the method of EVA3D \cite{hong2023evad}. They were trained on the deep fashion image dataset \cite{Liu_2016_CVPR}, along with the estimated SMPL \cite{loper2015smpl} model parameters and camera perspective. The model was trained 400,000 iterations with a learning rate of 0.002 and a batch size of 64. We adopt the Adam optimizer.

In the GAN inversion step, we first used a trained generator to generate 50,000 images by random sampling of latent codes under the condition that the camera Angle of view was positive and the human body attitude parameters were fixed, and recorded the correspondence between the images and latent codes. Then, we perform decoupling image encoding according to the method mentioned in the section \ref{sec:enc}. The CLIP model used for attribute decoupling is ViT-L/14. 

The predefined text library encompasses seven distinct attributes: gender, sleeve-length, top-color, top-type, pants-length, pants-color, and pants-type. The number of the attributes' category is decided by the segmentation performance. By merging the segmentation results of similar components such as arms and sleeves, seven distinct subgraph segments can be obtained. 

\subsection{Qualitative Results} 
Our generation result is shown as Figure \ref{fig:gen}. The results demonstrate that our approach can generate a 3D avatar matching the input text, incorporating various attributes of the human body. More vivid results can be found on the project page\footnote{project page: https://iecqgong.github.io/text2avatar/}. We compared our method with other text-to-3D baselines as Figure \ref{fig:com} shows. Results demonstrate that existing methods struggle to deal with coupled cues, while our approach achieves better decoupling of high-quality generation. It can be observed that Dreamfusion \cite{dreamfusion} has a missing upper part of the body, while 3DFuse \cite{3dfusion} has redundant human structures. Avatars generated by AvatarCLIP \cite{avatarclip} and Text2Avatar both have complete human structures and exhibit the best performance. Comparing AvatarCLIP and Text2Avatar more carefully, it can be seen that Text2Avatar can generate more accurate results (green box) and higher-resolution faces (red box) under coupled textual prompt.

\subsection{Quantitative Comparisons} 
We compare our method with existing text-to-3D methods\cite{dreamfusion,3dfusion,avatarclip}.
We highlight that these existing text-driven cross-modal methods are limited in their capability to handle coupled instructions.

\begin{figure}[tbp]
  \centering
  \mbox{} \hfill
  \hfill
  \includegraphics[width=1.0\linewidth]{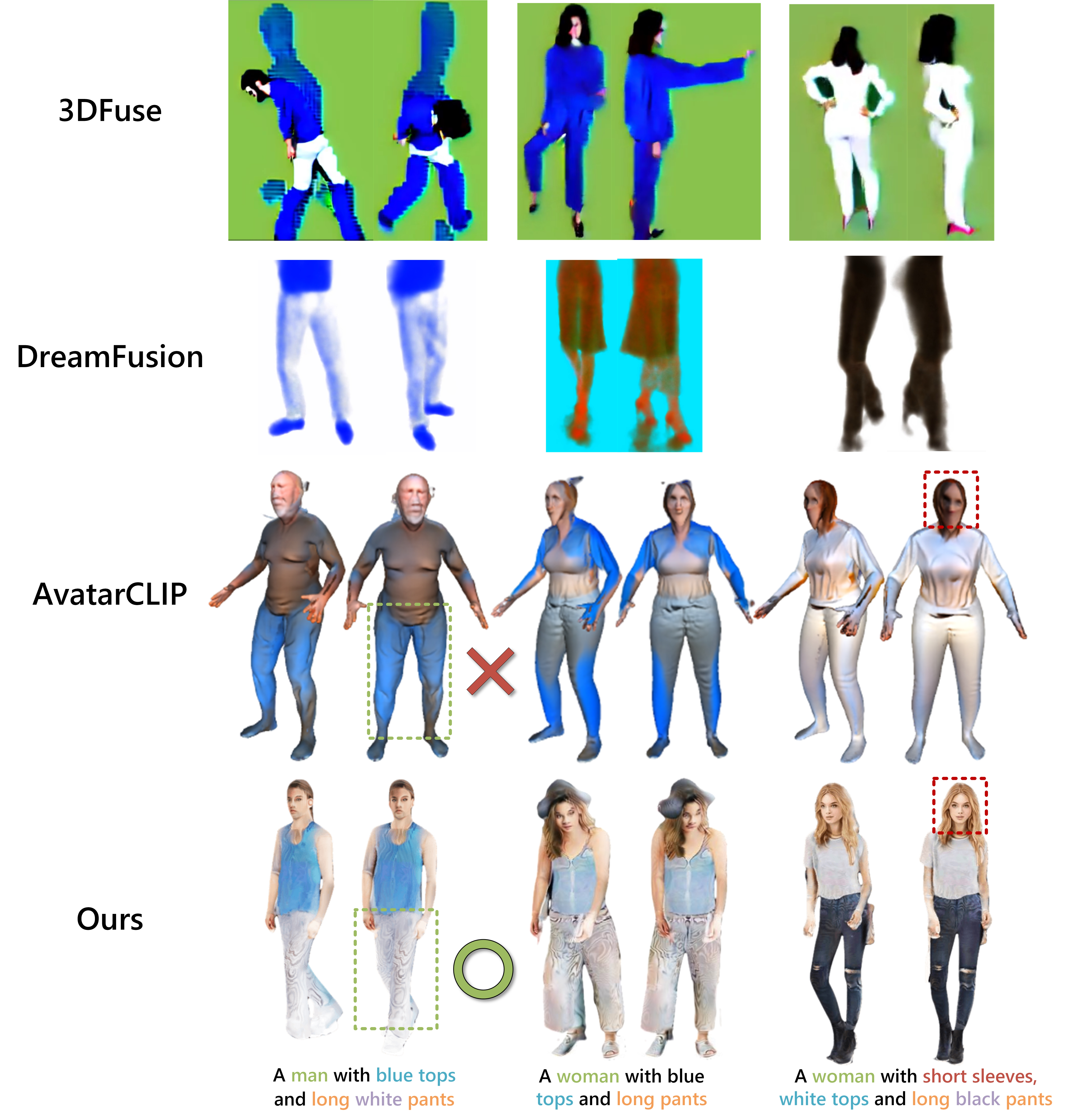}
  \hfill \mbox{}
  \vspace{-7mm}
  \caption{\label{fig:com}%
           Comparison between baselines and Text2Avatar. }
    \vspace{-5mm}
\end{figure}

\textbf{Attribute Accuracy.}
Due to the long time required and high cost involved of our baselines, we only allowed each model to generate 20 samples and manually compared their semantic matching accuracy as attribute accuracy. However, considering the obvious limitations of the comparative models in terms of visual results as Figure \ref{fig:com} shows, we believe that this number is sufficient to demonstrate the superiority of our model.

\textbf{R-Precision.}
We employ volumetric rendering to converted 3D samples to 2D. We utilize the CLIP model \cite{clip} to calculate the correlation between the images and the textual features, and take the average value as the R-Precision. Considering that the samples generated by the model are labeled as ``human", we include the match degree of the generated results with the word ``human" in the calculation of R-Precision.

As shown in Table \ref{table:com},
we omitted the attribute accuracy of some models for difficult-to-measure attributes, which are unable to discern accurately due to low generation quality, or for which the model did not generate the corresponding body part information (e.g., Dreamfusion \cite{dreamfusion} did not generate the upper body of the avatar).

The results demonstrate that our model exhibits significant superiority in attribute accuracy across various attributes, while also achieving the best R-Precision.

\begin{table}[h!]
  \begin{center}
  \vspace{-5mm}
    \caption{Ablation study result.w/o stands for without.\label{tab:abl}}
    \vspace{-3mm}
    \small 
    \setlength{\tabcolsep}{3pt} 
    \begin{tabular}{c|cc|cc} 
    \toprule
      \multirow{2}{*}{\textbf{Methods}} & \multicolumn{2}{c|}{\textbf{Attribute Accuracy}} & \multicolumn{2}{c}{\textbf{R-Precision}}\\
      & Pants-color & Sleeve-length & ViT-B/32 & ViT-L/14\\
      \hline
      w/o codebook & 0.55 & 1.00 & 77.71 & 82.65 \\
      w/o segmentation & 0.45 & 0.30 & 76.64 & 81.66\\
      origin & \textbf{0.80} & \textbf{1.00} & \textbf{78.52} & \textbf{83.30} \\
    \bottomrule
    \end{tabular}
  \end{center}
\vspace{-6mm}
\end{table}

\subsection{Ablation Studies}
To validate the effectiveness of our codebook design approach, we conducted ablative experiments by removing the codebook and segmentation module separately. In fact, the attributes that we define can be roughly divided into three categories: gender attribute, length attribute, and color attribute. As the accuracy of gender is consistently high, we have chosen the indicative length-related attribute (Sleeve-length) and the color-related attribute (Pants-color) to illustrate the superiority of our method, along with the R-Precision.

The experimental results are shown in Table \ref{tab:abl}, indicating that the segmentation operation and codebook significantly improve the recognition accuracy and R-precision. This is because segmentation converts local information of human attributes into global information, helping CLIP overcome the disadvantage of local information confusion, while codebook effectively increases the controllability between mappings.

\section{CONCLUSIONS}
We propose Text2Avatar, a method for generating realistic-style 3D Avatars from coupled multi-attribute description text. We highlight that the Multi-Modal Encoder module can serve as a plugin after retraining, therefore providing flexibility. 
In this way, different clothing of the human body can be easily obtained from textual information solely.
\section*{Acknowledgements}
This research was partly supported by Shenzhen Key Laboratory of next generation interactive media innovative technology (Grant No: ZDSYS20210623092001004), the China Postdoctoral Science Foundation (No.2023M731957), the National Natural Science Foundation of China under Grant 62306165, 62072242 and 62361166670.

\vfill\pagebreak

\bibliographystyle{IEEEbib}
\bibliography{ref}

\end{document}